\documentclass[runningheads]{llncs}
\usepackage{graphicx}
\usepackage{subfig}
\usepackage[export]{adjustbox}

\usepackage{rotating}
\usepackage{multirow}
\usepackage{array}
\usepackage{xcolor}
\usepackage{amssymb}
\usepackage{algpseudocode,algorithm,algorithmicx}  
\usepackage{pgf, tikz}
\usetikzlibrary{automata,positioning,shapes,arrows,bayesnet}

\begin{document}

\title{Paired-Consistency: An Example-Based Model-Agnostic Approach to Fairness Regularization in Machine Learning}

\titlerunning{Paired-Consistency}

\author{Yair Horesh \inst{1} \and Noa Haas\inst{1} \and Elhanan Mishraky \inst{1} \and Yehezkel S. Resheff \inst{1} \and Shir~Meir~Lador \inst{1} }
\authorrunning{Horesh et al.}
\institute{Intuit Inc.  \\
\email{\{first\_last\}@intuit.com}}
\newcommand{\floor}[1]{\lfloor #1 \rfloor}
\makeatletter
\newcommand*{\indep}{%
  \mathbin{%
    \mathpalette{\@indep}{}%
  }%
}
\newcommand*{\nindep}{%
  \mathbin{
    \mathpalette{\@indep}{/}%
  }%
}
\newcommand*{\@indep}[2]{%
  \sbox0{$#1\perp\m@th$}
  \sbox2{$#1=$}
  \sbox4{$#1\vcenter{}$}
  \rlap{\copy0}
  \dimen@=\dimexpr\ht2-\ht4-.2pt\relax
  \kern\dimen@
  \ifx\\#2\\%
  \else
    \hbox to \wd2{\hss$#1#2\m@th$\hss}%
    \kern-\wd2 %
  \fi
  \kern\dimen@
  \copy0 
}
\makeatother
\maketitle

\begin{abstract}
As AI systems develop in complexity it is becoming increasingly hard to ensure non-discrimination on the basis of protected attributes such as gender, age, and race. Many recent methods have been developed for dealing with this issue as long as the protected attribute is explicitly available for the algorithm. We address the setting where this is not the case (with either no explicit protected attribute, or a large set of them). Instead, we assume the existence of a fair domain expert capable of generating an extension to the labeled dataset - a small set of example pairs, each having a different value on a subset of protected variables, but judged to warrant a similar model response. We define a performance metric -  paired consistency. Paired consistency measures how close the output (assigned by a classifier or a regressor) is on these carefully selected pairs of examples for which fairness dictates identical decisions. In some cases consistency can be embedded within the loss function during optimization and serve as a fairness regularizer, and in others it is a tool for fair model selection. We demonstrate our method using the well studied \textit{Income Census} dataset. 
\end{abstract}


\section{Introduction}
The notion of fairness is deeply rooted in human kind \cite{fehr1999theory,kahneman1986fairness,forsythe1994fairness} and even in other intelligent species \cite{brauer2006apes,brosnan2003monkeys}. In practice, fairness is elusive. Due to the nature of complex systems we operate within, both conscious and unconscious cognitive biases, and lack of complete knowledge, it is extremely hard to guarantee fairness even in the presence of sufficient good will. 

When considering large scale machine learning systems we must proceed with caution. On the one hand the current trend of increased adoption of machine learning is a unique opportunity to clean the slate and utilize automation for objectivity and fairness. On the other hand, it is often unclear how to operate when labeled data is derived from historic processes with questionable fairness. Furthermore, models are typically optimized with respect to some measure of performance on the task they are designed to do, a process that has no relation to fairness and almost always favors the average outcomes.

Arguably, there are three levels of adherence to ethical and fair practice of AI. In the best case, a system designer is clearly and unequivocally fair. This option is unusual if at all possible (for instance because of the often mutually-exclusive notions of fairness). Failing that, fairness disputes could be based on the notion of what is fair. In this case there should be little or no doubt about the integrity and intentions of the designer. For instance, when trading off individual fairness in order to obtain better group fairness. Finally, a system can be outright discriminatory.

Too often systems already in use are discovered to be outright discriminatory. It has been claimed that the Correctional Offender Management Profiling for Alternative Sanctions (COMPAS) system -- a system widely used by courts for predicting a defendant's risk of recidivism within 2 years using 137 features, has low prediction power and a strong racial bias \cite{2018SciA....4O5580D,propublica}.

While explicit discrimination is easy to detect and remove, features that are correlated to discriminating attributes are much harder to detect. Consider a classification or a regression task via some method $f(\cdot)$, with a dataset at hand $\big\{(d^{(i)},x^{(i)},y^{(i)})\big\}_{i=1}^{N}$, where $Y$ is the target variable, and $D$ and $X$ are separate feature spaces, $D$ contains the protected variables. Restricting the fitting of $f(\cdot)$ solely through $X$ clearly eliminates explicit discrimination with respect to features in $D$. However, although $Y\indep D|X$, we might still observe association between the predictions and $D$, since $Y \nindep D$ in the general case, even when $D$ is not included in the fitting stage \cite{pedreshi2008discrimination}. Hence, even if the protected attributes \textit{is} completely removed from the input data, this does not guarantee fairness. 

What characterizes a fair system? Several competing notions of fairness have been recently proposed in the machine learning literature. The simplest notion of fairness is \textit{demographic parity} \cite{calders2009building} (Table \ref{tab:defs}). To satisfy demographic parity the underlying proportion of the protected variable should be preserved within the classification process (for example, an equal number of male and female candidates should pass the exam). This criteria is meant to preserve \textit{group fairness}, and as such has several drawbacks. 

First, demographic parity can be naively achieved through random assignment of the target $\hat{Y}$ in the underprivileged group according to the distribution in the privileged group. This clearly achieves demographic parity, but misses the original objective of achieving a useful prediction of $Y$ in the entire population. This sort of unfair selection in the underprivileged group leads to two detrimental consequences. Not only is it unfair to deserving individuals in the underprivileged group, since they have a smaller probability of being selected than similar individuals in the majority group, but this process also leads to a selection of less appropriate individuals from the underprivileged group, setting them up for failure, and reinforcing stigma. This behavior is prone to happen for instance when $D$ represents minority groups, and there is little or no training data available for one of the demographics \cite{hardt2016equality}.

Second, the demographic parity approach assumes that the population affiliation and the target are independent. For example, consider the case of a hiring process for a job associated primarily with males. It is possible that the few females that do submit an application are on average more qualified than the male applicants. In such a case, we would expect a fair process not to preserve the original proportions, and indeed accept a higher proportion of the female applicants. 

Metrics which asses how well a model maintain demographic parity are disparate impact and parity difference which are the ratio and difference between the conditional probability of the positive class given the binary protected variable, and \textit{prejudice index}, which quantifies the mutual information between the target $Y$ and $D$ \cite{kamishima2011fairness} (Table \ref{tab:defs}).


Other notions of fairness focus on the individual rather than on the group, and try to assure that an \textit{individual} is treated fairly irrespective of $D$. These measures are based on the notion that similar individuals should be treated similarly by the system. One approach for achieving individual fairness is to aim at removing the information about $D$ from the original data representation $X$. This however is not an easy thing to do, for instance, studies on gender bias show that removing bias from word embedding is extremely hard \cite{genderbiaswordembedings}. In another study it was shown that removing demographic information from representations of text is also not an easy task \cite{elazar2018adversarial}. Recently the adversarial learning framework has been suggested to reduce bias in learned representations for fair models \cite{Zhang:2018:MUB:3278721.3278779}, and in the context of fair and private recommendations \cite{resheff2018privacy}. 

Metrics for evaluating a system's individual fairness include average odds difference and equal opportunity difference \cite{hardt2016equality} which attempt to guarantee uniform aggregate behavior of the predictor with respect to a binary protected variable, and consistency, which compares a model\textsc{\char13}s classification prediction of a given data item x to its k-nearest neighbors \cite{zemel2013learning} (see table \ref{tab:defs} for a glossary of definitions and metrics in the fairness literature).

The aim of research on fair AI methods and metrics is to help design systems that adhere to the notion of fairness the system designer chooses to adopt. This is the \textit{how}, not the \textit{what}. Assuming good intentions, the normative question of what we consider to be fair is a separate issue that should be addressed by a much wider community. All of the above notions rely on the clear, categorical definition of the discriminatory variable $D$, and its presence in the data. In this paper we address the question of designing fair models with any specific notion of individual fairness. 

Unlike the methods described above, we deal in a setting where there is no simple categorical variable (or small number thereof) that defines demographic groups for which fairness should be guaranteed. This can be the case when there are \textit{no} explicit protected attributes (but still discrimination danger from demographic-correlated attributes). Alternatively there may be \textit{many} protected attributes, with either many categories or continuous values. In these cases the existing methods do not apply or are infeasible to use. The proposed method uses an expert to mine a small auxiliary dataset that refines the desired notion of fairness, a performance metric, and a regularization method to obtain it. 

The core idea is that if we had many pairs with similar \textit{merit}, in the sense that with respect to the chosen notion of fairness they should get the same treatment, we could force the model to behave that way. To this end we assume the existence of a (fair) domain expert who is able to label pairs of instances. Each pair has a different value in a subset of protected variables $D$, and the expert asserts that a fair model should output a similar response for them. Note that this methodology doesn't require a defined similarity metric between samples, a property which allows more flexibility in applying it.

The proposed method is general in two aspects. While previous fairness inducing methods require explicit access to the protected attribute $D$, the pairs generated by the expert in our method can be selected on the basis of any implicit idea of a potentially discriminatory variable, that doesn't have to be directly measurable. The second aspect of generality is the wide applicability to machine learning models. We suggest a tree-training variant, and a gradient-descent training variant, together covering the majority of widely used machine learning methods. 

The rest of the paper is structured as follows: in the next section we describe the proposed method of paired-consistency for fairness. Next, we describe experiments and results on the "Census Income" dataset, comparing the proposed method to alternatives that do require direct access to the restricted variables, using a large set of fairness criteria. We finish with a discussion on the pros and cons of the proposed method and future direction of research.

\section{Methods}
In this section we present the method of paired-consistency. The setting we operate in is a dataset: $\big\{(d^{(i)},x^{(i)},y^{(i)})\big\}_{i=1}^{N}$, where each example consists of features $x$, an additional (and possibly empty) set of explicitly given restricted variables $d$, and a target $y$. In addition, access to a fair domain expert is assumed. The expert may be a literal human expert, or an algorithm or method used as a surrogate. 

The fair domain expert comes equipped with a notion of fairness, and of the potential attributes that must be protected from discrimination. These attributes may be explicit (i.e. contained in $d$ -- for example gender or age), or more complex constructs that the expert is able to determine based on a sample $(d, x, y)$ (for example being of an underprivileged background). Upon request, the expert returns pairs:

$$\big\{(x_1^{(j)}, x_2^{(j)})\big\}_{j=1}^{M}$$

\noindent denoted as the \textit{consistency set}. Each of these pairs consists of the features from two examples of the original dataset, that obey two requirements. First, the pair represents two examples which are different with respect to the protected attribute or construct, as determined by the expert. Second, based on the remainder of the information the expert is able to judge that the two samples warrant a similar response by the model. 

An interesting property of the paired-consistency method is that it is able to protect from discrimination not only in the absence of the explicit protected variable, but even when the construct of interest is not directly measurable, as long as the fair domain expert is able to match pairs which are different with respect to it. Cases when this may be of use include when individuals with certain special circumstances are historically under-represented and a fair selection process might therefore attempt to take this into consideration. Furthermore, this method is able to mix and combine fairness with respect to different potential sources of discrimination, by combining the sets of pairs derived from each one. 

In this setting, we suggest a simple model-agnostic method which can be used to assess (if used for model selection) or help assure fairness (if used as a regularizer) of a classification or regression model for predicting $y$. We define a paired-consistency score, which measures how similar is an output (in terms of assigned class, or predicted score) a model produces for paired members, for classification:

\begin{equation}
     \frac{1}{M} \sum_{j=1}^{M}I[\hat{y}_1^{(j)} = \hat{y}_2^{(j)}]
\label{eq:consistency-classification}
\end{equation}

\noindent where $\hat{y}_1^{(j)} = f(x_{1}^{(j)})$ is the model output, and $I[\cdot]$ is the indicator function. This measures the fraction of the pairs on which the model agrees. Likewise for regression:

\begin{equation}
     1 - \frac{1}{M \cdot \delta_{max}} \sum_{j=1}^{M}(\hat{y}_1^{(j)} - \hat{y}_2^{(j)})^2
\label{eq:consistency-regression}
\end{equation}

\noindent where $\delta_{max}$ is the maximal square difference, used to normalize the measure into $[0,1]$ (this is necessary only when comparing models, otherwise the measure to minimize becomes $\frac{1}{M} \sum_{j=1}^{M}(\hat{y}_1^{(j)} - \hat{y}_2^{(j)})^2$). 

The consistency score is embedded within the loss function as a fairness regularization term, to make the model consistency aware. This is done by adding the measure (eq. \ref{eq:consistency-regression}) to the objective, multiplied by a trade-off parameter to determine the relative importance of the main objective and the paired-consistency. Any algorithm trained via gradient-descent (and variants) can be adapted to incorporate this additional loss component. In addition, we suggest a variant for training of trees, where the local optimization criterion is augmented in a similar way to include the fraction of pairs kept intact in each split. Results of both these types are presented below (Section \ref{sec:results}).  

In addition (or alternatively for fairness-based model selection), the score is calculated \textit{post-hoc}, and can be aggregated with other performance metrics, or used as part of a performance-fairness trade-off. A good classifier will be accurate but also consistent in the scoring of the pairs. To this end we define the \textit{PRC score} as the weighted Harmonic Mean of Precision, Recall, and Paired-Consistency, which as an F1-score analogue it is a natural candidate for integrating consistency in evaluation of models. 

The proposed method also allows a natural integration of the certainty of the expert. Suppose that together with the pairs, the expert also produces a weight reflecting how sure they are that the pair is indeed a fairness-match -- different on some subset of the protect variables or constructs, and deserving of the same treatment -- so that the expert now outputs $\big\{(x_1^{(j)}, x_2^{(j)}, w^{(j)})\big\}_{j=1}^{M}$. The classification paired-fairness measure (Equation \ref{eq:consistency-classification}) will thus become: 

\begin{equation}
     \frac{\sum_{j=1}^{M} w^{(j)} \cdot I[\hat{y}_1^{(j)} = \hat{y}_2^{(j)}]}{M \cdot \sum_{j=1}^{M}w^{(j)} } 
\label{eq:consistency-classification-w}
\end{equation}

\begin{table}
    \centering
\renewcommand{\arraystretch}{1.5}
\begin{tabular}{ll}
\hline
Name & Definition \\
\hline
demographic parity & $Pr(\hat{Y} = 1 | D = 0) = Pr(\hat{Y} = 1 | D = 1)$  \\
parity difference  &  $Pr(\hat{Y} = 1 | D = 0) - Pr(\hat{Y} = 1 | D = 1)$  \\
disparate impact   &  $\frac{Pr(\hat{Y} = 1 | D = 0)}{Pr(\hat{Y} = 1 | D = 1)}$   \\ 
equalized odds \cite{hardt2016equality} & $ TPR_{D = 0} = TPR_{D = 1} ; FPR_{D = 0} = FPR_{D = 1}$ \\ 
average odds difference &  $\frac{1}{2}\left[(FPR_{D = 0} - FPR_{D = 1})
   + (TPR_{D = 0} - TPR_{D = 1}))\right]$  \\
equal opportunity \cite{hardt2016equality} & $ TPR_{D = 0} = TPR_{D = 1} $ \\
equal opportunity differene  &  $ TPR_{D = 0} - TPR_{D = 1} $ \\ 
consistency \cite{zemel2013learning}& $1 - \frac{1}{N} \sum_{i=1}^{N} |\hat{y}_i - \frac{1}{k} \sum_{j \in knn(i)}
\hat{y}_j |$ \\
prejudice index \cite{kamishima2011fairness}& $\sum \hat{p}(y, d) ln \frac{\hat{p}(y, d)}{\hat{p}(y)\hat{p}(d)}$ \\
paired-consistency (ours) & $\frac{1}{M} \sum_{j=1}^{M}I[\hat{y}_1^{(j)} = \hat{y}_2^{(j)}]$ \\ 
\hline 

\end{tabular}

\caption{Glossary: several common definitions and measures in the fairness literature. (F/TPR - false/true positive rate, D - protected variable, Y - the target, $\hat{Y}$ - model prediction, knn - k nearest neighbors). }
\label{tab:defs}
\end{table}

\section{Results}
\label{sec:results}

We demonstrate the paired-consistency method using a well-known dataset, and both for tree-based and gradient-based model training. The dataset we use to demonstrate the method is the "Census Income" dataset \cite{Dua:2019,kohavi1996scaling} derived from the 1994 census in the US. 
In this dataset the set of discriminatory variables $D$ appears explicitly in the data. We proceed to generate consistency pairs in order to emulate the case where an expert is called upon to generate pairs in the absence of explicit information. In some of the experiments below we also leave the protected attributes in the set of features used by the model in order to test the effect of paired-consistency regularization on the utilization of restricted information by the model. This setting emulates the standard case where the outright discriminatory features are indeed excluded from the model (gender, age, race, etc.), but other highly correlated features are included. 


\begin{figure}
    \centering
    \subfloat[\label{left} Income gender bias]{{\includegraphics[width=5.7cm,valign=t]{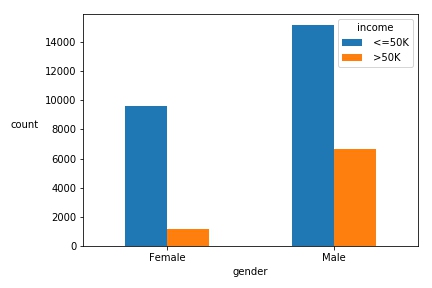}}}%
    \qquad
    \subfloat[\label{right} Work hours and income bias]{{\includegraphics[width=5.7cm,valign=t]{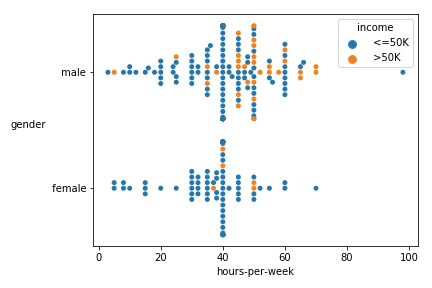} }}%
    \caption{The gender-income bias and interaction with number of weekly work hours as reflected in the \textit{census income} dataset.}%
    \label{fig:census_gender_income}%
\end{figure}

The "Census Income" dataset was chosen because it contains a combination of attributes a person is born with or has no control over (gender, race, age, native-country) and attributes that reflect their preferences and decisions in life (occupation, weekly work hours, education, martial status). The Census Income dataset contains data pertaining to over 32,000 individuals. The dataset also contains a binary field indicating high income (over $50K$ as of the 1994 census) which will be used as the target variable for our tests. The overall fraction of high-income individuals in the dataset is $26\%$.

In this experiment we model the likelihood of being a high-income individual based on individual traits. The setting we have in mind is one where the output of a model of this sort will impact the way people are treated. As such, we would need to take care not to discriminate on the bases of certain properties of individuals. While it is clear that if left untreated, attributes such race, gender and native-country will have predictive power for income level, some attributes are both relevant and at the same time a proxy of discriminating attributes. Figure \ref{fig:census_gender_income} illustrates this ambiguity. The gender income imbalance reflected in the census data is captured by Figure \ref{fig:census_gender_income}\subref{left}. However, the swarm-plot figure (Figure \ref{fig:census_gender_income}\subref{right}) shows there is a gender difference in the probability of being in the high-income group, and also a difference in distribution of weekly work hours. As a result, the otherwise innocent (and intuitively relevant) variable of number of work hours, gives away some information about the protected variable (which is gender). 

In order to test the effect of paired-consistency on the fairness and performance of a predictive model, we use a decision tree and a Logistic Regression model. The motivation for using these simple models is that the readily interpretable outcome lets us better understand the effect of the fairness regularization, and at the same time these are representatives of the two major classes of machine learning models at this time (i.e. tree based and gradient based training). Paired-consistency is added to the Logistic Regression model by inserting the mean square deviation in output among pairs  (Equation \ref{eq:consistency-regression}) directly to the loss function, via a trade-off parameter. 

For tree training, we add the fairness metric as an extension to the Gini Index used in the tree creation. In order to adapt the measure (Equation \ref{eq:consistency-classification}) to the local criterion of tree growing, for a given split we seek to maximize the number of pairs that go in the same direction. To this end we add to the Gini Index a term that is the percent of the pairs arriving at the node that are kept intact following the split (i.e. both examples in the pair go to the same side). This term is multiplied by a trade-off parameter that controls the relative importance of the fairness regularization in the tree construction. As expected when training trees, this is a local optimization criterion. Feature importance in the resulting model was measured using the column permutation method (and using the \textit{eli5} Python package \cite{eli5}). Experiments were conducted using a 80-20 train-test split, making sure original pairs are kept in the same set. All categorical features are encoded as dummies, leading to a total of 58 features. For regularization, and to ensure an interpretable result, we limit decision tree depth to 5 and the minimal items in a leaf to 5.

\subsubsection{Baseline}
\begin{figure}
\centering{}
\includegraphics[width=11cm]{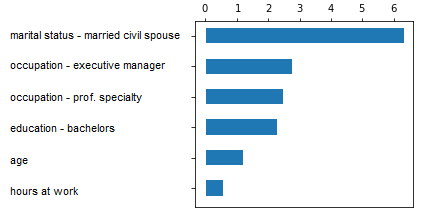}
\caption{Top-$6$ feature importance in the basic tree model for \textit{census income} (arbitrary units).} 
\label{fig:census_weight}
\end{figure}

The tree baseline consists of a regular decision tree, achieving an accuracy level of 82.6\%. Figure \ref{fig:census_weight} shows the most important features resulting from this model. Interestingly the list is topped by the "married civil spouse" indicator, followed by occupation and education indicators. We choose to focus on the next most important variable -- age, which we chose as the discriminating factor we want to mitigate in the following experiments with consistency pairs. It is important to note that in a real use-case the protected variables would undoubtedly be removed from the model themselves, but this blindness would not suffice to achieve fairness because of correlated features that remain in the input. We keep the age variable in the model in order to emulate the case of additional information correlated with the protected variable(s). The Logistic Regression baseline achieves an accuracy level of $82.9\%$.

\subsubsection{Consistency pairs}
We automatically created 3,062 consistency pairs by selecting pairs of individuals who are similar in all other aspects except for age (while ignoring income). We picked pairs where the age gap is of 10 years or more. Obviously this is a toy example, and in a more complicated model a fair domain expert will likely be necessary for this purpose. 

To minimize the paired-consistency fairness penalty the model must predict for each pair the same predicted income and avoid discriminating on the basis of age. We test the model with four levels of the fairness-regularization trade-off parameter. Tables \ref{tab:feature-importance} and \ref{tab:num-pairs-importance} summarizes the results of this experiment. Table \ref{tab:feature-importance} shows that the importance rank of the variable under consideration is reduced monotonically as the weight of the fairness-regularization is increased, from rank $5$ in the naive model to rank $10$ with larger weights, reaching a plateau with a weight of $1$. The effect flattens out when the percent of pairs classified together is $100\%$ after which the fairness penalty is $0$ and therefore further increasing the weight is irrelevant. Overall accuracy of the model is virtually unaffected, and even slightly improved together with increased fairness. Table \ref{tab:num-pairs-importance} shows this effect -- as the number of pairs used for the consistency regularization increases, the importance of the age feature in the model decreases. We note that even a relatively small number of pairs ($500$ pairs, versus the $32,561$ examples in the dataset) was sufficient to significantly mitigate the age bias, although the change obtained in importance of the restricted variable from rank-$5$ to rank $7$ may not be sufficient. 

\begin{table}[]
\renewcommand{\arraystretch}{1.2}
    \centering

    \begin{tabular}{|llll|}
    \hline
    Regularization weight & Age importance (rank) & \% Accuracy & \% Pairs intact\\
    \hline
    0 & 1.35 (5) & 82.9 & 97.6\\
    0.1 & 0.2 (8) & 83.0 & 99.8\\
    1 & 0.03 (10) & 83.0 & 100\\
    10 & 0.03 (10) & 83.8 & 100\\
    \hline
    \end{tabular}

\caption{Effect of fairness regularization on importance rank of and score of the protected variable, model accuracy, and percent of consistency pairs labeled consistently.}
\label{tab:feature-importance}
\end{table}

\begin{table}[]
\renewcommand{\arraystretch}{1.2}
    \centering

    \begin{tabular}{|llll|}
    \hline
    \# Pairs & Age importance (rank) & \% Accuracy & \% Pairs intact\\
    \hline
    100 & 1.50 (5) & 82.7 & 100\\
    500 & 0.58 (7) & 82.7 & 100\\
    1000 & 0.33 (7) & 82.8 & 100\\
    3062 & 0.03 (10) & 83.0 & 100\\
    \hline
    \end{tabular}

\caption{Effect of the number of consistency pairs on importance rank and score of the protected variable, model accuracy, and percent of consistency pairs labeled consistently. The fairness-regularization weight (eta) is fixed to 0.5.}
\label{tab:num-pairs-importance}
\end{table}

\begin{figure}[th]
\centering{}
\includegraphics[width=12.0cm]{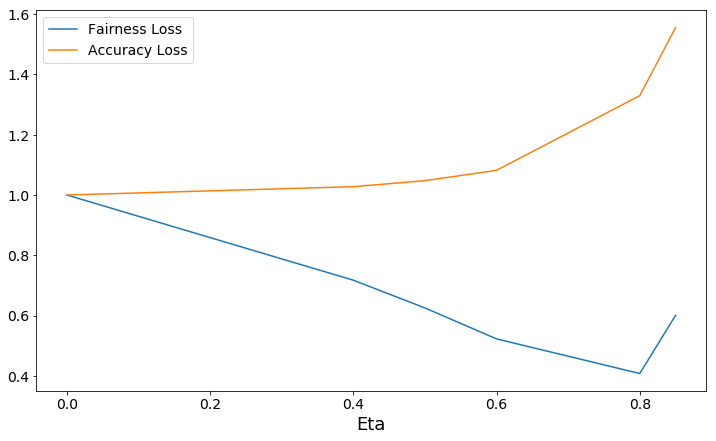}
\caption{Trade-off of Logistic Regression loss and paired-consistency loss components as a function of the trade-off parameter eta. Values are normalized and presented as fraction of the respective loss components for $\eta = 0$.}
\label{fig:training-process}
\end{figure}

Logistic Regression with paired-consistency regularization displays the expected trade-off between overall model accuracy and fairness (as measured by paired-consistency score). Figure \ref{fig:training-process} shows this trade-off between the two components of the loss function (accuracy and fairness components). Increasing the trade-off parameter $\eta$ shifts the weight towards the fairness objective, and in turn leads to a decrease in the paired-consistency component of the objective, together with an increase in the original Logistic Regression loss component. The effect of the trade-off parameter is shown as the fractional change in each of the loss components, compared to the baseline ($\eta = 0$; regular Logistic Regression). Results indicate that for the price of a modest decline in accuracy, the fairness component of the loss can be reduced by as much as half. This favorable trade-off can be seen in the relative slopes of the lines in Figure \ref{fig:training-process} for eta in the range of $0-0.5$.  

We further compare the fairness-accuracy trade-off in various methods for predicting the target variable using the dataset. The compared methods include both classification methods without any fairness regularization or constraints, and methodologies with the objective of a creating a fair model. We applied our paired-consistency regularization to Logistic Regression and decision tree models, and a prejudice remover \cite{kamishima2011fairness} regularization (with $\eta=10$) to a Logistic Regression model (using the \textit{AI Fairness 360} \cite{bellamy2018ai} python package). Methods were compared both in accuracy and in fairness. Accuracy was used to measure utility, and the metrics in Table \ref{tab:defs}, as well as our paired consistency metric were used to measure various aspects of model fairness.

Such comparisons provides a clear understanding with regard to the potential trade-off between predictive modeling utility and fairness. The ultimate result would be to achieve high fairness, while maintaining the model performance with respect to the prediction objective. As theoretically expected, almost all fairness-regularization methods hindered the model's accuracy, but this effect is not dramatic for the tested dataset and methods. All fairness regularization methods show improvements in the fairness metrics over the respective baselines. The consistency-paired regularized tree seems like the fairest model available here (both in group and individual fairness), but under-performs in terms of accuracy. Another aspect to learn from these results is the differentiation between the various fairness metrics -- while the overall trend is to improve fairness for all methods, each methodology outperform in a different fairness metric. This is of course expected. Each method optimizes with respect to a specific notion of fairness, and in general is likely therefore to do best on that, while being less competitive on others (this is especially the case when considering contradictory metrics such as group vs. individual fairness, where it is not possible generally to be optimal in both). 

Finally, we test the effect of the number of consistency pairs available on the accuracy-fairness trade-off when used as a fairness regularization in Logistic Regression. The importance of this stems from the high cost and effort involved in generating fairness pairs in some real-world situations. Unlike in our current experiment, it is not always possible to generate this auxiliary data automatically, and instead a human expert is used to label data pairs. Ideally, we would want to know how many pairs are necessary, and how the number of pairs is likely to impact the fairness measures.

Results are summarized in Table \ref{tab:comparison-num-pairs}. Logistic Regression models are trained with paired-consistency regularization ($\eta=0.4$). As expected, as the number of consistency-pairs used is increased, the paired-consistency score increases as well, from $0.705$ with $100$ pairs, to $0.945$ with $1000$ pairs. However, $500$ pairs are enough to obtain a score of $0.912$, not significantly less than the optimal value. A similar picture is reflected also in the other fairness measures. The accuracy of the fairness-regularized Logistic Regression models is seen to increase overall with the number of pairs. This can be explained by the arbitrary effect each pair has on the overall the model, which may average out when many pairs are applied. An analytic understanding of this phenomenon will be necessary in future research, as well as testing on additional datasets.

\begin{table}[bh]
\renewcommand{\arraystretch}{1.2}
    \centering

\begin{tabular}{|l|p{1.0cm}p{1.0cm}|p{1.2cm}p{1.2cm}p{1.5cm}p{1.5cm}|}
\hline
{} &   Tree &  Tree +pairs & LR &  LR\cite{kamishima2011fairness} (prej.)&  LR+pairs eta=0.4 &  LR+pairs eta=0.5 \\
\hline
Paired consistency score                               &  0.976 &                               1.000 &  0.912 &                 0.913 &             0.927 &              0.932 \\
Classification accuracy                            &  0.832 &                               0.834 &  0.829 &                 0.826 &             0.824 &              0.815 \\
Disparate impact                                   &  0.250 &                               0.299 &  0.193 &                 0.213 &             0.212 &              0.225 \\
Statistical parity diff.                      & -0.164 &                              -0.160 & -0.314 &                -0.314 &            -0.304 &             -0.318 \\
Equal opportunity diff.                       & -0.149 &                              -0.091 & -0.248 &                -0.227 &            -0.234 &             -0.220 \\
Average odds diff.                          & -0.103 &                              -0.073 & -0.215 &                -0.205 &            -0.204 &             -0.207 \\
\hline
\end{tabular}

\caption{Census Income: Comparison of different methodologies (with and without fairness constraints), in terms of performance and various fairness metrics. LR -- Logistic Regression. (prej.) -- Prejudice Remover (Table \ref{tab:defs}). tree/LR+pairs -- tree/LR with our method of paired-consistency.}
    \label{tab:comparison-methods-census}
\end{table}

\begin{table}[th]
\renewcommand{\arraystretch}{1.2}
    \centering

    \begin{tabular}{|p{1.5cm}|p{1.5cm}p{2.0cm}p{2.0cm}p{2.0cm}|}
    \hline
    \# Pairs  &  Accuracy & Paired \hspace{1cm} consistency & Parity \hspace{1cm} difference  &  Average odds \hspace{1cm} difference\\
    \hline
    100 & 0.759 & 0.705 & -0.490 & -0.477\\
    500 & 0.804 & 0.912 & -0.376 &-0.269\\
    1000 & 0.797 & 0.945 & -0.322 & -0.191\\
    3062 & 0.824 & 0.927 & -0.304 & -0.204\\
    \hline
    \end{tabular}

\caption{Logistic Regression with paired-consistency regularization ($\eta=0.4$). Effect of the number of consistency pairs on various performance and fairness scores.}
    \label{tab:comparison-num-pairs}
\end{table}

\section{Discussion and Conclusion}

With the rise in popularity of AI across many domains, questions of ethical use and fairness are revisited with renewed vigor. Many methods have previously been proposed to help create fair machine learning algorithms, when the variable which leads to potential discrimination is explicitly available in the data. However, this is not always the case. 

We present a simple yet powerful method to help mitigate discrimination in machine learning models, without using an explicit partitioning of the data with respect to a protected variable. Our approach relies on the ability of a fair domain expert to generate a set of pairs of examples which are equivalent based on all attributes except for a subset of the protected variables. We then assert that a fair model should treat the two examples in each of these pairs in a similar way, and define a measure of consistency that when added to a models loss function helps enforce fairness via consistency. This fairness regularization technique is shown to reduce the extent to which a decision tree model uses a forbidden feature in classification, and has favorable outcomes also with respect to other measures of fairness. 

The proposed method of paired-consistency is related to several existing techniques. In a broad sense, statistically defined group fairness methods try to ensure that on the group level the minority group is treated by the machine learning algorithm in a similar way to the majority group. In that case our method is similar by reducing the size of the groups to $1$ and having many such pairs. 

Experiments on a well-studied dataset in the fair model literature demonstrates the viability of our method, both for tree-based and gradient-based training (with Logistic Regression). It is interesting to note that in our experiments we see little to no decline in accuracy as we increasingly enforce fairness by adjusting the trade off parameter in the loss function. This sort of degenerate trade-off is likely not the general case. When the trade-off takes on a more substantial form, the additional paired-consistency term added to the loss function can be seen as a regularization mechanism against bias, in as much as it restricts the search during optimization to regions of higher fairness. The result in the general case will inevitably be some loss in overall model accuracy. 

One of the merits of the proposed approach is that it enables domain experts to take part in the fairness efforts and mitigate discrimination without the need to understand how the model works or what features or information it is based on. Even better, by using examples, the expert bypasses the need to formalize the sometimes elusive notion of fairness. In fact, the fairness labelling can and should be done by an expert prior to and independently from machine learning work. Since the labeling is independent from the methods used to make predictions, it can be seen as an extension of labeling for supervised learning rather than of the process of evaluating results of a model. However, new consistency pairs can be generated after a model is created to further evaluate its fairness properties with respect to discriminating variable of interest. Adding new consistency pairs doesn't require the model itself to be changed, only to be re-trained.  

There are several limitations to paired-consistency. First, our method assumes the existence and availability of a fair domain expert capable of generating consistency pairs. In many cases this is not a huge stretch, but other times may be infeasible due to time, cost, or trust. The current experiments do however suggest that a relatively small number of pairs (several hundred) are already sufficient to achieve most of the benefit. 

Even when such an expert is available, there is still a problem in agreeing on the notion of fairness to be used in the pairing of examples. This final issue is a fundamental problem of fair machine learning (and possibly fairness in general). The inability to agree on what is fair is inherently a limitation on the ability to design fair systems. We note that this question is outside of the scope of machine learning and engineering in general, and is rather a question to be tackled by a wider community in the broad spectrum of the humanities and social sciences.  

Future work will focus on extending the ideas brought here to other types of models and domains, and provide theoretical performance guarantees. In the case of deep learning (or more generally gradient optimized models), it is relatively straight forward to add the consistency term as specified in Equation \ref{eq:consistency-regression}, as is demonstrated for the Logistic Regression case here. It remains to be seen how this will effect different types of models, and how the effect varies between datasets. For other types of models it may be more difficult to incorporate this penalty, but the method can still be used independently for the purpose of fair model selection (since in that case it is a \textit{post-hoc} calculation on the model output, that doesn't depend on the model's cost function itself). 

An additional question of interest that we do not cover here is the effect of the number of consistency pairs necessary to achieve the goal. While we were able to generate a relatively large number for the current datasets (since it was an automatic process), in the more general case a human expert will manually pick consistency pairs, a potentially costly and time consuming effort. The cost and effort required limit the number of pairs that are feasible for any given dataset. 

Finally, we note that this method can be applied synergistically with other fairness techniques for model construction and evaluation. For example, this method can be used in conjunction with adversarial techniques for removing unwanted information from deep learning representations \cite{Zhang:2018:MUB:3278721.3278779,elazar2018adversarial} to obtain a model that is based on a representation devoid of the information about the protected variables on the one hand, and that enforces fairness in the notion of consistency on the other hand. Future work will focus on the interplay between the various trade-off parameters that emerge from such a construction. Similarly, it is interesting to investigate the effect of the pair sampling done by the fair domain expert, and cases when specific types of sampling lead to paired-consistency converging back to one of the previously proposed statistical notions of fairness.

\bibliographystyle{splncs04}
\bibliography{bib.bib}

\end{document}